%% file: DeepLiDARFlow.tex
\documentclass[letterpaper, 10 pt, conference]{ieeeconf}  % Comment this line out if you need a4paper

\IEEEoverridecommandlockouts                              % This command is only needed if 
\newcommand{\name}{\mbox{DeepLiDARFlow}} 
\title{\LARGE \bf
\name{}: A Deep Learning Architecture For Scene Flow Estimation Using Monocular Camera and Sparse LiDAR
}

\author{Rishav$^{*,1,2}$ \hspace{5mm} Ramy Battrawy$^{*,1}$ \hspace{5mm} Ren{\'e} Schuster$^1$ \hspace{5mm} Oliver Wasenm{\"u}ller$^{1,3}$ \hspace{5mm} Didier Stricker$^{1,4}$% <-this % stops a space
\thanks{$^{*}$Equal contribution}%
\thanks{$^{1}$German Research Center for Artificial Intelligence - DFKI, 
	Kaisers-\hspace*{3mm} lautern, 
	Germany:
	{\tt\small firstname.lastname@dfki.de}}% <-this % stops a space
\thanks{$^{2}$Birla Institute of Technology and Science - BITS Pilani, 
	Pilani, 
	India:
	\hspace*{3mm} {\tt\small f2016108@pilani.bits-pilani.ac.in}}
\thanks{$^{3}$University of Applied Sciences Mannheim, 
	Mannheim, 
	Germany:
	\hspace*{3mm} {\tt\small o.wasenmueller@hs-mannheim.de}}
\thanks{$^{4}$University of Kaiserslautern - TUK, 
	Kaiserslautern, 
	Germany}}
\usepackage{hyperref}
\hypersetup{
    linkcolor=blue,
	filecolor=magenta,      
	urlcolor=cyan,
}
\usepackage{graphicx}
\usepackage{float}
\usepackage{subcaption}
\usepackage{rotating}
\usepackage{multirow}
\usepackage{mwe}
\usepackage{xcolor}
\usepackage{amssymb}% http://ctan.org/pkg/amssymb
\usepackage{pifont}% http://ctan.org/pkg/pifont
\newcommand{\cmark}{\ding{51}}%
\newcommand{\xmark}{\ding{55}}%
\usepackage{tikz}
\usepackage{dashbox}
\usepackage{makecell}
\usepackage{url}
\usepackage{xspace}

\begin{document}

\newcommand*{\eg}{e.g.\@\xspace}
\newcommand*{\ie}{i.e.\@\xspace}

\newcommand\Tstrut{\rule{0pt}{2.6ex}}         % = `top' strut
\newcommand\Bstrut{\rule[-0.8ex]{0pt}{0pt}}

\maketitle
\thispagestyle{empty}
\pagestyle{empty}

%%%%%%%%%%%%%%%%%%%%%%%%%%%%%%%%%%%%%%%%%%%%%%%%%%%%%%%%%%%%%%%%%%%%%%%%%%%%%%%%

%abstarct
\input{abstract}

%%%%%%%%%%%%%%%%%%%%%%%%%%%%%%%%%%%%%%%%%%%%%%%%%%%%%%%%%%%%%%%%%%%%%%%%%%%%%%%%

%Introduction
\input{introduction}

\input{related_work}

\input{method}

\input{experiments}

\section{Conclusion}
In this paper, we presented our \name{} -- a novel deep learning architecture which takes monocular images and the corresponding sparse LiDAR measurements as input, employs a multi-scale late fusion of LiDAR and RGB features, and predicts dense scene flow.
In critical regions which contain difficulties like reflective surfaces, ill conditioned environment, shadows, and more, our \name{} shows superior performance over image-only methods. Moreover, we provided a robust localization compared to an image-only approach as well as a conventional approach. Compared to a LiDAR-only approach, we achieved a superior accuracy for scene flow estimation.
Our method obtained competitive performance on the challenging KITTI and FlyingThings3D data sets with very sparse LiDAR input ($< 1000$ points) and almost constant accuracy with different levels of input density.

\section*{Acknowledgment}
This work was partially funded by the Federal Ministry of Education and Research Germany in the project VIDETE (01IW18002).

%\addtolength{\textheight}{-15cm}	% This command serves to balance the column lengths
\IEEEtriggeratref{26}

%%%%%%%%%%%%%%%%%%%%%%%%%%%%%%%%%%%%%%%%%%%%%%%%%%%%%%%%%%%%%%%%%%%%%%%%%%%%%%%%

%%%%%%%%%%%%%%%%%%%%%%%%%%%%%%%%%%%%%%%%%%%%%%%%%%%%%%%%%%%%%%%%%%%%%%%%%%%%%%%%

%%%%%%%%%%%%%%%%%%%%%%%%%%%%%%%%%%%%%%%%%%%%%%%%%%%%%%%%%%%%%%%%%%%%%%%%%%%%%%%%

%%%%%%%%%%%%%%%%%%%%%%%%%%%%%%%%%%%%%%%%%%%%%%%%%%%%%%%%%%%%%%%%%%%%%%%%%%%%%%%%

\bibliographystyle{IEEEtrans}
\bibliography{bibliographyfile}

\end{document}

%% file: abstract.tex
\begin{abstract}
Scene flow is the dense 3D reconstruction of motion and geometry of a scene. Most state-of-the-art methods use a pair of stereo images as input for full scene reconstruction.
These methods depend a lot on the quality of the RGB images and perform poorly in regions with reflective objects, shadows, ill-conditioned light environment and so on. 
LiDAR measurements are much less sensitive to the aforementioned conditions but LiDAR features are in general unsuitable for matching tasks due to their sparse nature.
Hence, using both LiDAR and RGB can potentially overcome the individual disadvantages of each sensor by mutual improvement and yield robust features which can improve the matching process.
In this paper, we present \name{}, a novel deep learning architecture which fuses high level RGB and LiDAR features at multiple scales in a monocular setup to predict dense scene flow. Its performance is much better in the critical regions where image-only and LiDAR-only methods are inaccurate. 
We verify our \name{} using the established data sets KITTI and FlyingThings3D and we show strong robustness compared to several state-of-the-art methods which used other input modalities. 
The code of our paper is available at \url{https://github.com/dfki-av/DeepLiDARFlow}.
\end{abstract}

%% file: introduction.tex
\section{Introduction}
Robust understanding about 3D geometry and dynamic changes in the environment is very important for autonomous vehicles, robot navigation, advanced driver assistance systems and so on.
In this context, scene flow estimation is an essential task which aims to the reconstruct 3D geometry as well as 3D motion of each observed point in the entire scene.
Hence, dense scene flow enriches the perceptual information which makes it very useful to increase the reliability of autonomous systems.  

Most of the popular scene flow methods use stereo images.
But there is an inherent disadvantage with image-based methods because they depend completely on the quality of the image.
Therefore, scene flow estimation gets extremely difficult if the images contain insufficient details in some regions due to reflective surfaces, shadows, bad illumination and many more.

LiDAR sensors are much less sensitive to the aforementioned environmental conditions.
Thus, they can possibly act as anchor points in the regions where the RGB features are not robust.
A problem with LiDAR sensors is that they get expensive as the density of points they provide increases. 
Hence an ideal method should be able to work with very sparse LiDAR measurements in order to ensure its cost effectiveness.
The fusion of high level features of RGB images and robust features of LiDAR measurements for dense matching can potentially result in highly accurate scene flow estimates even under bad environmental conditions.
Recently, Battrawy et al. \cite{battrawy2019lidar} proposed a conventional approach that fuses LiDAR measurements into stereo images for dense scene flow estimation. 
They show impressive improvement compared to a stereo-only setup, however, their approach is computationally inefficient. 
Thus, we propose an end-to-end learning-based approach to estimate dense scene flow from sparse LiDAR and RGB images (see Fig. \ref{teaser}). To the best of our knowledge, our \name{} is the first approach that uses the fusion of sparse LiDAR measurements and RGB images in a monocular setup for dense scene flow estimation.

\input{Graphics/teaser.tex}

Our \name{} learns high level features of sparse LiDAR measurements and RGB images at multiple scales and fuses them into each other in an end-to-end learning-based fashion.
It aims to resolve critical regions of bad illumination, shadows, reflective objects in RGB image and produce robust features for matching.
Overall, the main contributions of this work are:
\begin{itemize}
	\item A novel deep learning strategy for dense scene flow estimation by fusing sparse LiDAR and RGB images in a monocular setup.
	\item A novel multi-scale fusion strategy of RGB and LiDAR features for dense scene flow estimation.
	\item Exhaustive experiments, showing the superior performance of our \name{} over image-based methods in critical regions with reflective objects, bad illumination, etc.
	\item Overall competitive and robust results against different state-of-the-art algorithms which use other input modalities.
\end{itemize}

%% file: Graphics/teaser.tex
\begin{figure}[t]
	\centering
	\begin{subfigure}[c]{0.02\linewidth}
		\rotatebox[origin=c]{90}{\textit{\scriptsize Input}}
	\end{subfigure}
    \begin{subfigure}[c]{0.46\linewidth}
        \includegraphics[width=\linewidth]{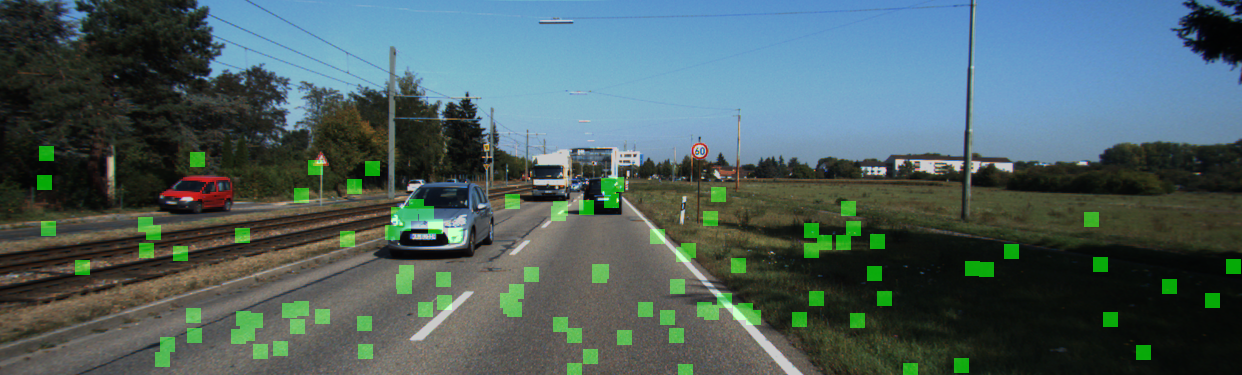}
    \end{subfigure}
    \hspace{0.01\linewidth}
    \begin{subfigure}[c]{0.46\linewidth}
        \includegraphics[width=\linewidth]{Graphics/Teaser_images/000039_mask_overlay.png}
    \end{subfigure} 
    \vspace{3 mm}
    
	\begin{subfigure}[c]{0.02\linewidth}
		\rotatebox[origin=c]{90}{\textit{\scriptsize D0}}
	\end{subfigure}
    \begin{subfigure}[c]{0.46\linewidth}
        \includegraphics[width=\linewidth]{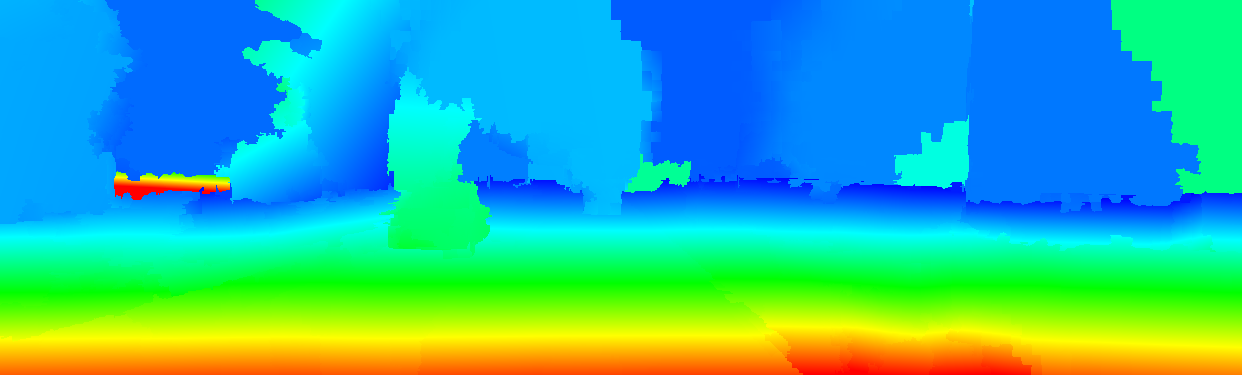}
    \end{subfigure}
    \hspace{0.01\linewidth}
    \begin{subfigure}[c]{0.46\linewidth}
        \includegraphics[width=\linewidth]{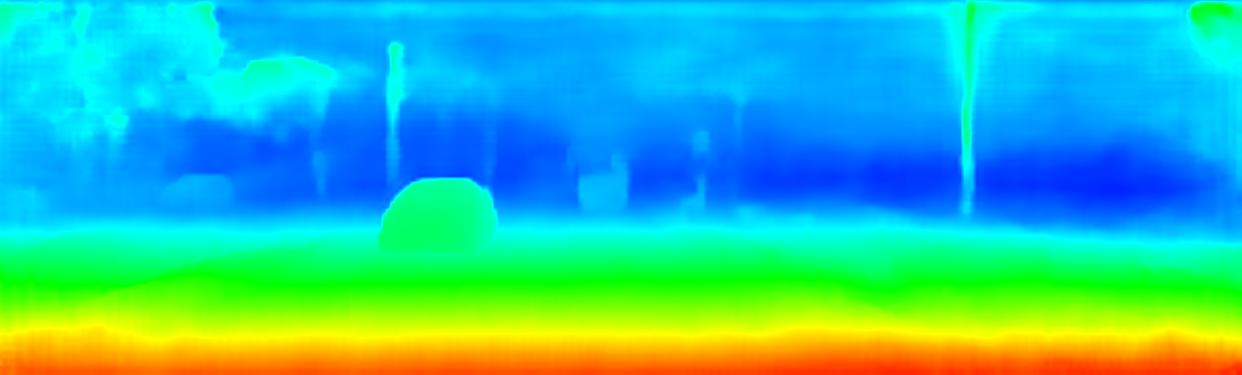}
    \end{subfigure} 
    \vspace{3 mm}
    
	\begin{subfigure}[c]{0.02\linewidth}
		\rotatebox[origin=c]{90}{\textit{\scriptsize Fl}}
	\end{subfigure}
    \begin{subfigure}[c]{0.46\linewidth}
        \includegraphics[width=\linewidth]{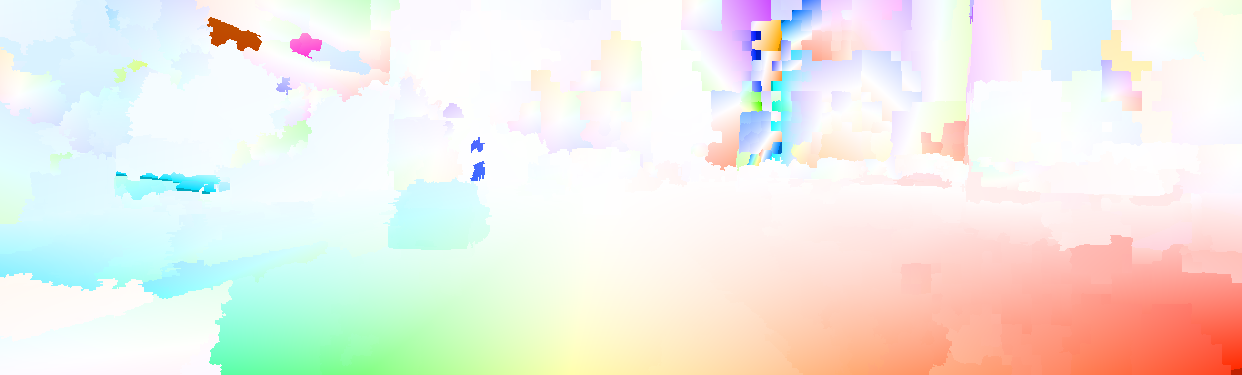}
    \end{subfigure}
    \hspace{0.01\linewidth}
    \begin{subfigure}[c]{0.46\linewidth}
        \includegraphics[width=\linewidth]{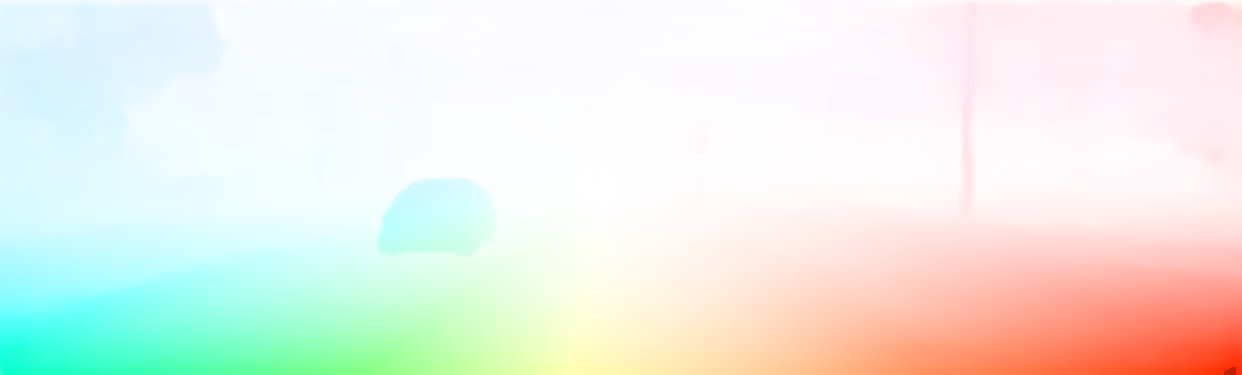}
    \end{subfigure}
    \vspace{3 mm}
    
	\begin{subfigure}[c]{0.02\linewidth}
		\rotatebox[origin=c]{90}{\textit{\scriptsize Error Map}}
	\end{subfigure}
    \begin{subfigure}[c]{0.46\linewidth}
        \includegraphics[width=\linewidth]{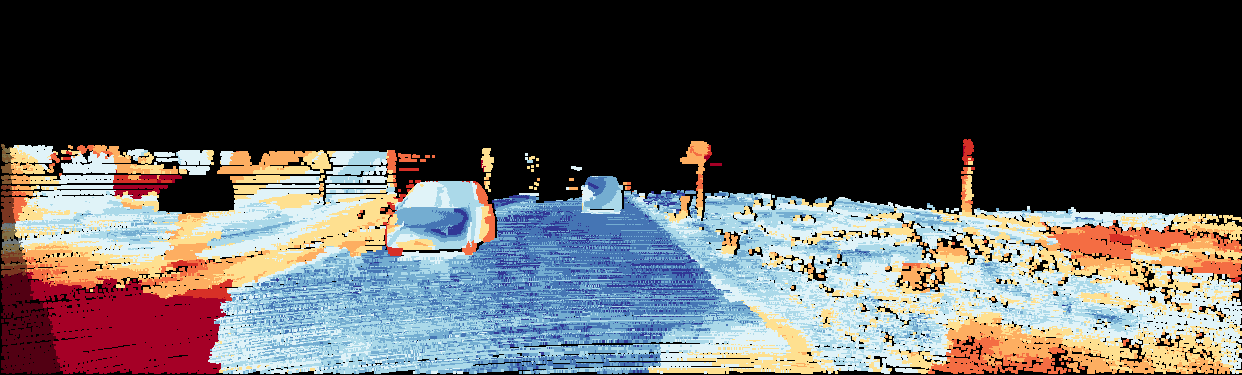}
    \end{subfigure}
    \hspace{0.01\linewidth}
    \begin{subfigure}[c]{0.46\linewidth}
        \includegraphics[width=\linewidth]{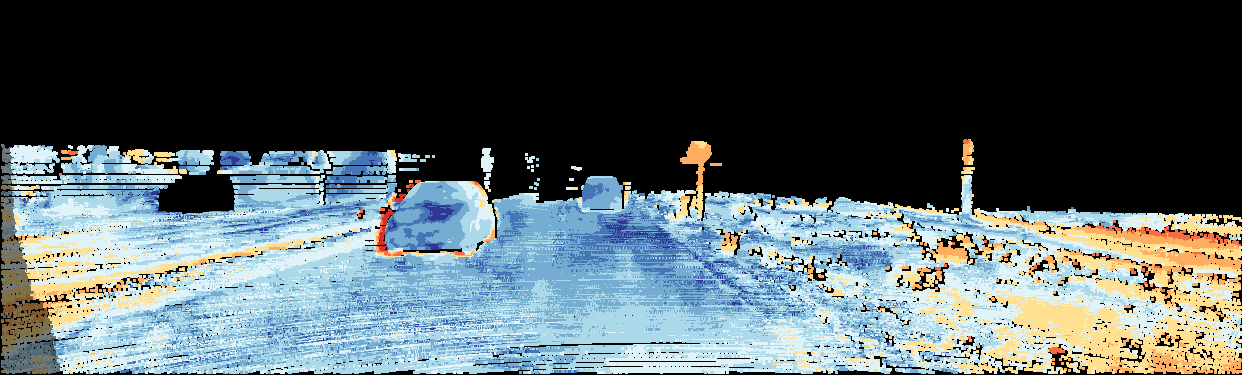}
    \end{subfigure}   
    \vspace{0.5 mm}
    
    \hspace{0.03\linewidth}
    \begin{subfigure}[c]{0.46\linewidth}
  		\centering
  		\small Conventional approach \cite{battrawy2019lidar}
        
    \end{subfigure}
    \hspace{0.01\linewidth}
    \begin{subfigure}[c]{0.46\linewidth}
  		\centering
        \small Our \name{}
    \end{subfigure}
    
    \caption{We introduce our \name{}, a novel deep learning architecture which fuses a monocular image and the corresponding sparse LiDAR measurements (shown as green spots on input image) for dense scene flow estimation. For very sparse LiDAR ($\sim$100 points), our \name{} outperforms comfortably the conventional scene flow approach (monocular version of \cite{battrawy2019lidar}) which employs such a fusion.}
	\label{teaser}
\end{figure}

%% file: related_work.tex
\section{Related Work}

	\subsection{Image-based Scene Flow}
	Most of the image-based scene flow methods utilize a pair of stereo images at two time steps, like the early variational methods \cite{vcech2011scene, huguet2007variational, isard2006dense, li2008multi}.
	The improvements brought about by deep convolutional neural networks (CNNs) for various computer vision tasks \cite{he2016deep, krizhevsky2012imagenet} over traditional approaches, were successfully transferred to dense pixel-wise matching tasks.
	FlowNet \cite{dosovitskiy2015flownet} is the first deep learning method developed to predict dense optical flow.
	SceneFlowNet \cite{ilg2018occlusions} is the first to use an end-to-end deep learning approach for scene flow using a pair of stereo images.
	Recently, PWOC-3D \cite{saxena2019pwoc}, DWARF \cite{aleotti2020learning} and SENSE \cite{jiang2019sense} propose light weight end-to-end networks which operate with the stereo image setup.
	DRISF \cite{ma2019deep} applies piece-wise rigid planes model \cite{vogel2013piecewise} and employs a combination of deep learning and conventional approaches.
	
	As alternative to the stereo setup, some methods use RGB-D cameras for dense scene flow estimation \cite{gottfried2011computing, herbst2013rgb, quiroga2014dense}.
	Qiao et al. \cite{qiao2018sf} are the first to develop a deep learning method that utilizes  RGB-D images for scene flow estimation.
	However, RGB-D setup performs poorly for outdoor scene flow estimations due to sensor range limitations. 
	Approaches like \cite{teng2018occlusion, yang2018every, yin2018geonet, zou2018df} use the power of multi-task CNNs by poising scene flow estimation from monocular images as a problem of single view depth and optical flow estimation.
	Mono-SF \cite{brickwedde2019mono} is a recent method that jointly estimates the 3D structure and motion of the scene by combining multi-view geometry and single-view depth information.
	Unlike most of these methods, our \name{} solves the scene flow problem as a whole in an end-to-end fashion.

	The major problem with purely image-based approaches is their heavy reliance on the image quality.
	These methods usually perform poorly in critical image regions with poor illumination, shadows or reflective objects.
	These are the regions where robust measurements from a LiDAR sensor are extremely useful.
	Our \name{} takes the advantage of these measurements and fuses them into the image domain to improve scene flow estimates. 

	\subsection{LiDAR-only Scene Flow}
	Some methods use point clouds to directly estimate scene flow. 
	In this context, FlowNet3D \cite{liu2019flownet3d} is among the first to propose a neural network architecture which utilize point clouds only.
	PointFlowNet \cite{behl2019pointflownet} uses a highly compartmentalized architecture to estimate scene flow from point clouds.
	HPLFlowNet \cite{gu2019hplflownet} takes inspiration from Bilateral Convolutional Layers (BCL) \cite{gadde2016superpixel} that restore structural information from unstructured point clouds.
	The two major problems with LiDAR-only approaches are the difficulty of matching unstructured data and the inherent low resolution compared to cameras.
	Our \name{} overcomes the individual disadvantages of each sensor by mutual improvement, hence proposing a novel sensor setup with strong potential for robust and accurate dense scene flow predictions.

	\subsection{LiDAR and Image-based Scene Flow}
	Recently, scene flow estimation in a heterogeneous sensor environment was proposed by LiDAR-Flow \cite{battrawy2019lidar}.
	However, this work focuses on considerable dense scene flow improvement over stereo-only approach by using a pair of stereo images and LiDAR measurements.
	Different from LiDAR-Flow, our \name{} aims to resolve the stereo camera dependency entirely by the fusion of a monocular camera and a LiDAR sensor.
	This is a much more challenging task, because the LiDAR information can not just be used to resolve ambiguous image cues, but is the only source of 3D information of the scene.
	Therefore to obtain a dense scene flow result, the sparse LiDAR measurements need to be converted into a dense representation.
	To the best of our knowledge, \name{} is the first approach that explores to this sensor with monocular setup for dense scene flow.

%% file: method.tex
\input{method_figures.tex}

\section{Method}

For scene flow estimation, the input of our \name{} is RGB images ($I^t$, $I^{t+1}$) and the corresponding LiDAR measurements ($D^t$, $D^{t+1}$) at two consecutive time steps $t$ and $t+1$ respectively. Our \name{} fuses the high level features of $I^t$, $I^{t+1}$ and $D^t$, $D^{t+1}$ to predict dense scene flow through three main modules: The feature extraction module, the fusion module, and the scene flow module. The following sections describe each module in details.

\subsection{Feature Extraction Module} \label{feature_extraction_module}
The feature extraction module consists of four multi scale feature pyramid networks for $I^t$, $I^{t+1}$, $D^t$, and $D^{t+1}$.
PWOC-3D \cite{saxena2019pwoc} uses a feature pyramid network to extract features with strong semantics and localization at multiple scales. Having features at multiple scales helps in tackling problems like large motion for dense pixel-wise matching.
The pyramids of RGB and LiDAR input are similar in structure to the feature pyramid network in PWOC-3D.
However, feature extraction from LiDAR data differs in the set of operations and layers we use.
In \cite{uhrig2017sparsity} it was shown that regular convolution fails to perform equivalently with varying density or pattern of sparse input.
Therefore, sparsity-aware convolution is proposed, which uses a binary sparsity mask for normalization.
As a further development, Eldesokey et al. \cite{eldesokey2019confidence} propose a confidence convolution which uses differentiable confidence volumes to indicate sparsity and the reliability of the densification.
We use the same concept of confidence convolution \cite{eldesokey2019confidence}, max-confidence pooling (for down-sampling), and nearest neighbor up-sampling to account for the sparse nature of LiDAR measurements during feature extraction.
The resolution of features is halved at each level of the pyramid and each level consists of two convolutions.
All pyramids have 6 levels, hence the final map is of $\frac{1}{64}$ resolution of the original input.
Afterwards, the features are successively decoded and up-sampled until $\frac{1}{4}$ of the input resolution is reached again.
The final features at a particular level $l$ are denoted by $i^t_l$, $i^{t+1}_l$, $d^t_l$, and $d^{t+1}_l$ for RGB and LiDAR input at the two time steps respectively. 
Feature pyramids for either the two RGB images or the two LiDAR measurements share their weights.
Fig. \ref{figure:deeplidarflow} presents more details of the feature extraction module.

\subsection{Fusion Module} \label{fusion_module}
\begin{figure}[t]
    \centering
    \includegraphics[width=\linewidth,keepaspectratio]{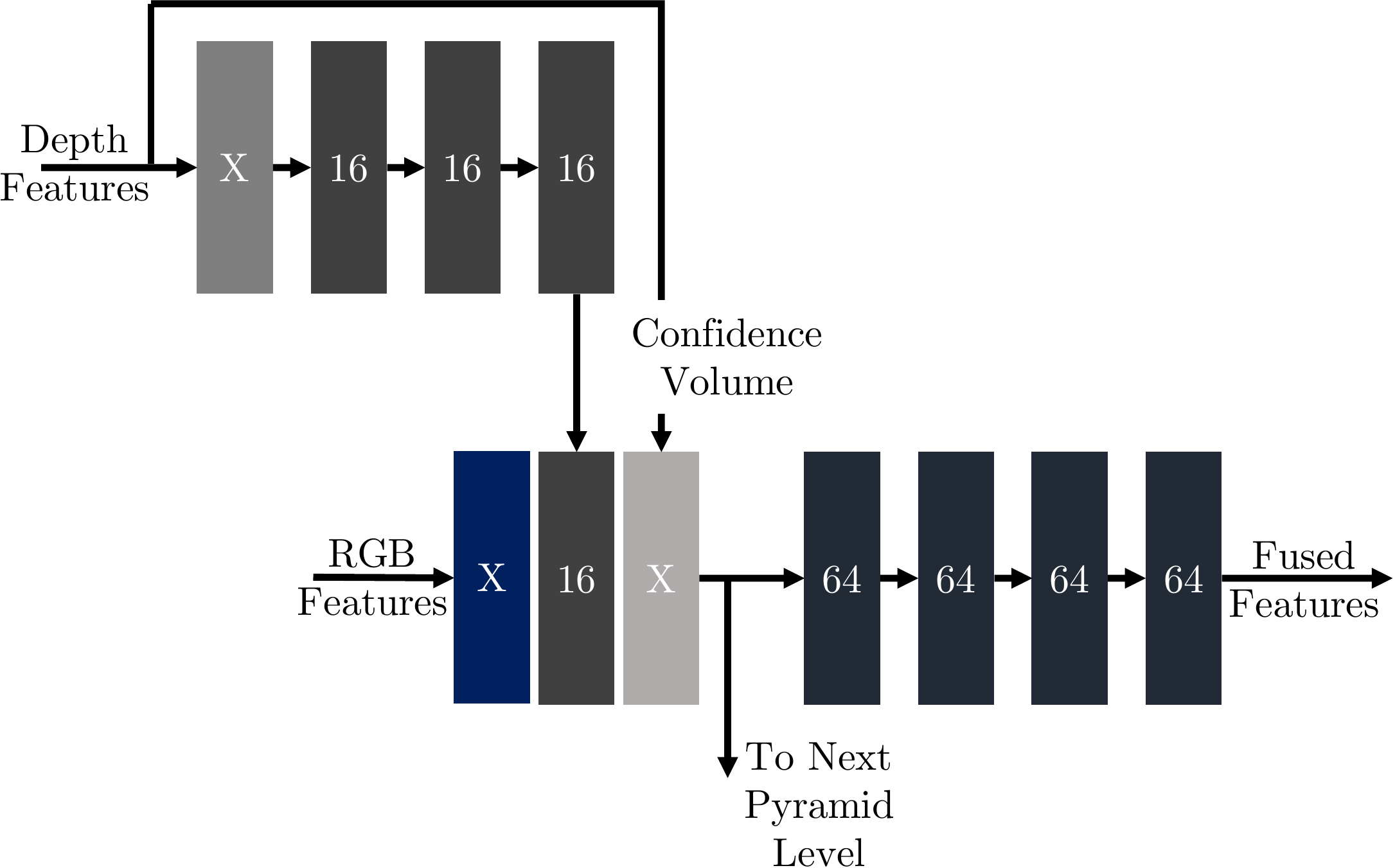}
    \caption{The fusion module at a particular scale level. The depth features obtained from the confidence pyramid go through a series of convolutions as a preprocessing step before finally being used for fusion. $X$ denotes the number of channels in the feature pyramid for that specific level. The fused features go through a series of convolutions to ensure proper mixing of the two heterogeneous information.}
    \label{figure:fusion}
\end{figure}

The fusion of heterogeneous RGB and depth information is an important part of our approach.
On the one hand, the depth information from the LiDAR feature is supposed to refine the image features to improve dense matching.
On the other hand, dense RGB information is used to guide the densification of the sparse LiDAR measurements to obtain a dense depth representation.
Previous work \cite{eitel2015multimodal, eldesokey2019confidence} experimented with early and late fusion, of which the late fusion strategy was shown to perform better.
Our \name{} builds on this finding and extends the late fusion of high level RGB and LiDAR features into a multi scale late fusion and prediction strategy.
With increasing level $l$, $d^t_l$ (and $d^{t+1}_l$) get more and more dense and semantically strong, but there is only little structural information depending on the density of the LiDAR input.
The RGB features $i^t_l$ and $i^{t+1}_l$ are rich in structural information.
The fusion module is responsible for the combination of structured RGB and unstructured LiDAR features to produce high level features for matching.
These features combine the robustness and accuracy of LiDAR measurements and the rich textural and structural information from RGB images.
The features at a particular level $l$ ($i^t_l$ and $d^t_l$, same for $i^{t+1}_l$ and $d^{t+1}_l$) are concatenated along the channel dimension into a feature volume which then goes through several convolutions while maintaining the input spatial dimensions (see Fig. \ref{figure:fusion}).
The fusion is performed for levels $l=6$ till $l=2$, \ie there is continuous fusion during the top down branch of the feature pyramids.
The fusion modules for the two time steps $t$ and $t+1$ share their weights.
The fused features at a level $l$ are denoted as $f^t_l$ and $f^{t+1}_l$ (the features of the reference frame and the next frame respectively).
These features are then sent to the scene flow module for final prediction of dense scene flow on each scale.

To give meta-guidance to the fusion module, the confidence volumes of the LiDAR features are concatenated with the (raw) RGB features before fusion (see Fig. \ref{figure:deeplidarflow}), at each scale ($l=6$ to $l=2$).
Since the confidence is a reliability measure of the depth features, this way, the fusion module is more flexible in how the two heterogeneous feature maps are fused.
The improvement brought about the concatenation of confidence maps is proved with the help of an ablation study discussed in Section \ref{design_decisions}.

\subsection{Scene Flow Module} \label{scene_flow_module}
\begin{figure}[t]
	\centering
	\includegraphics[width=0.9\linewidth,keepaspectratio]{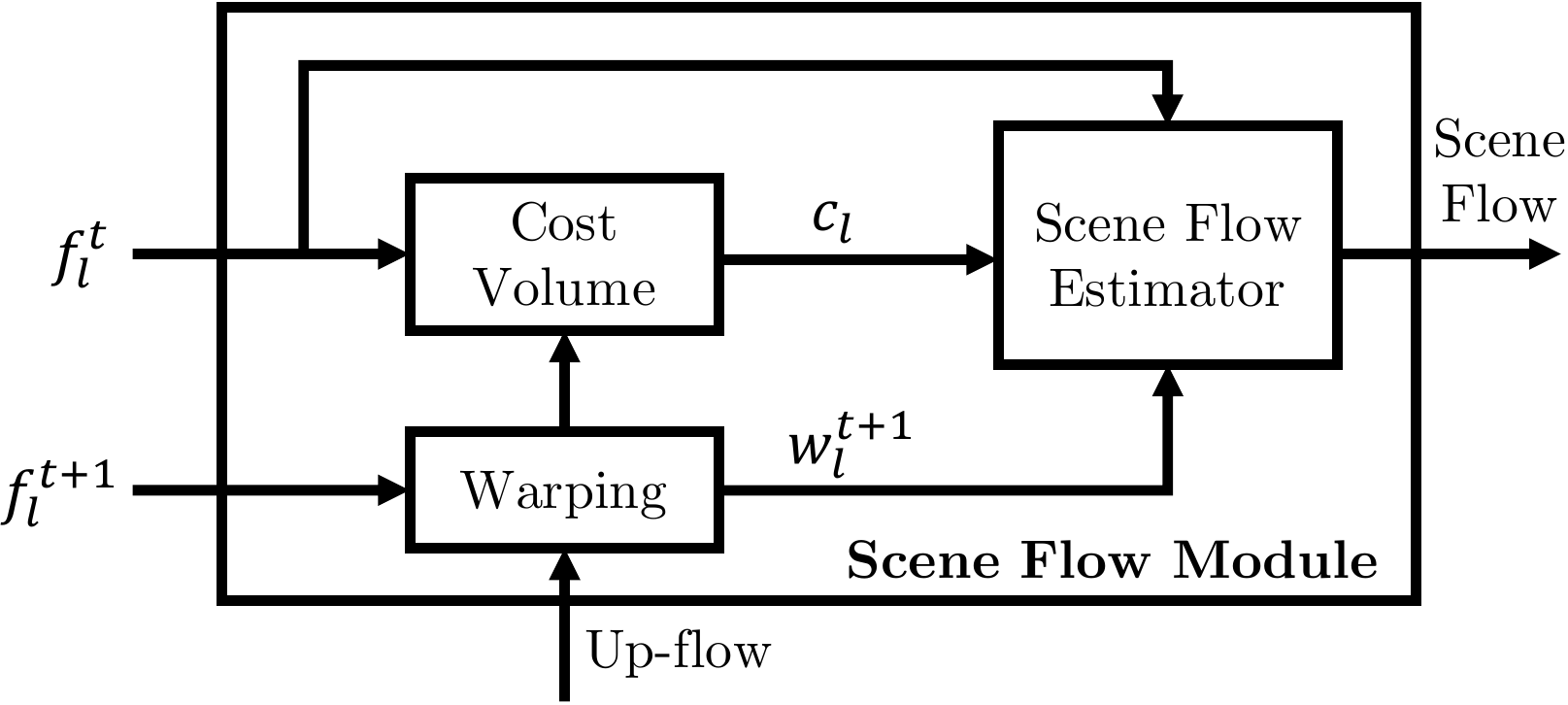}
	\caption{The scene flow module. $f^{t}_{l}$ ,$f^{t+1}_{l}$ are the fused features of the frames at time $t$ and $t+1$. Up-flow denotes the optical flow estimate from the previous level. This module further comprises of the warping layer, the cost volume layer and the scene flow estimator. The output of the last level is further refined by a context network as in \cite{saxena2019pwoc}.}
	\label{figure:sceneflow}
\end{figure}

The scene flow module (Fig. \ref{figure:sceneflow}) is the final component of \name{}.
At any particular level $l$ ($l=2$ to $l=6$), it comprises of a warping layer, a cost volume layer, and the scene flow estimator.
The blocks mentioned above differ from the ones used in PWOC-3D \cite{saxena2019pwoc} in the following aspects.
The input to this module are $f^t_l$ and $f^{t+1}_l$, i.e., the output features from the respective fusion modules.
Only a single 2-dimensional warping operation is needed, which warps $f^{t+1}_l$ towards $f^t_l$ to form $w^{t+1}_l$.
%The warped features of $f^{t+1}_l$ and $f^t_l$ are concatenated and given as input to the occlusion network which predicts a soft occlusion mask for $f^{t+1}_l$. The mask is used in the same way as PWOC-3D to mask out the occluded pixels from $f^{t+1}_l$ (the features after masking is denoted as ${f_m}^{t+1}_l$).
$w^{t+1}_l$ and $f^t_l$ are fed to the cost volume layer, which computes a 2D cost volume (denoted as $c_l$) in the same way as PWOC-3D \cite{saxena2019pwoc}.
$c_l$, $w^{t+1}_l$, and $f^t_l$ are then concatenated and given as input to the scene flow estimator which predicts the final, dense 4D scene flow at level $l$.
When the final level $l=2$ is reached, the dense prediction is further refined with a residual prediction from the context network. The context network gets $f^{t+1}_l$, $f^{t}_l$ and the last level features from the scene flow estimator as input. Fig. \ref{figure:sceneflow} gives a schematic view of the entire module.
Note that at $l=6$, \ie the lowest resolution, there is no previous flow estimate.
Instead the initial flow is assumed to be zero, resulting in no warping, \ie $w_6^{t+1} = f_6^{t+1}$.

%% file: method_figures.tex
\begin{figure*}[t]
    \centering
    \includegraphics[width=\linewidth]{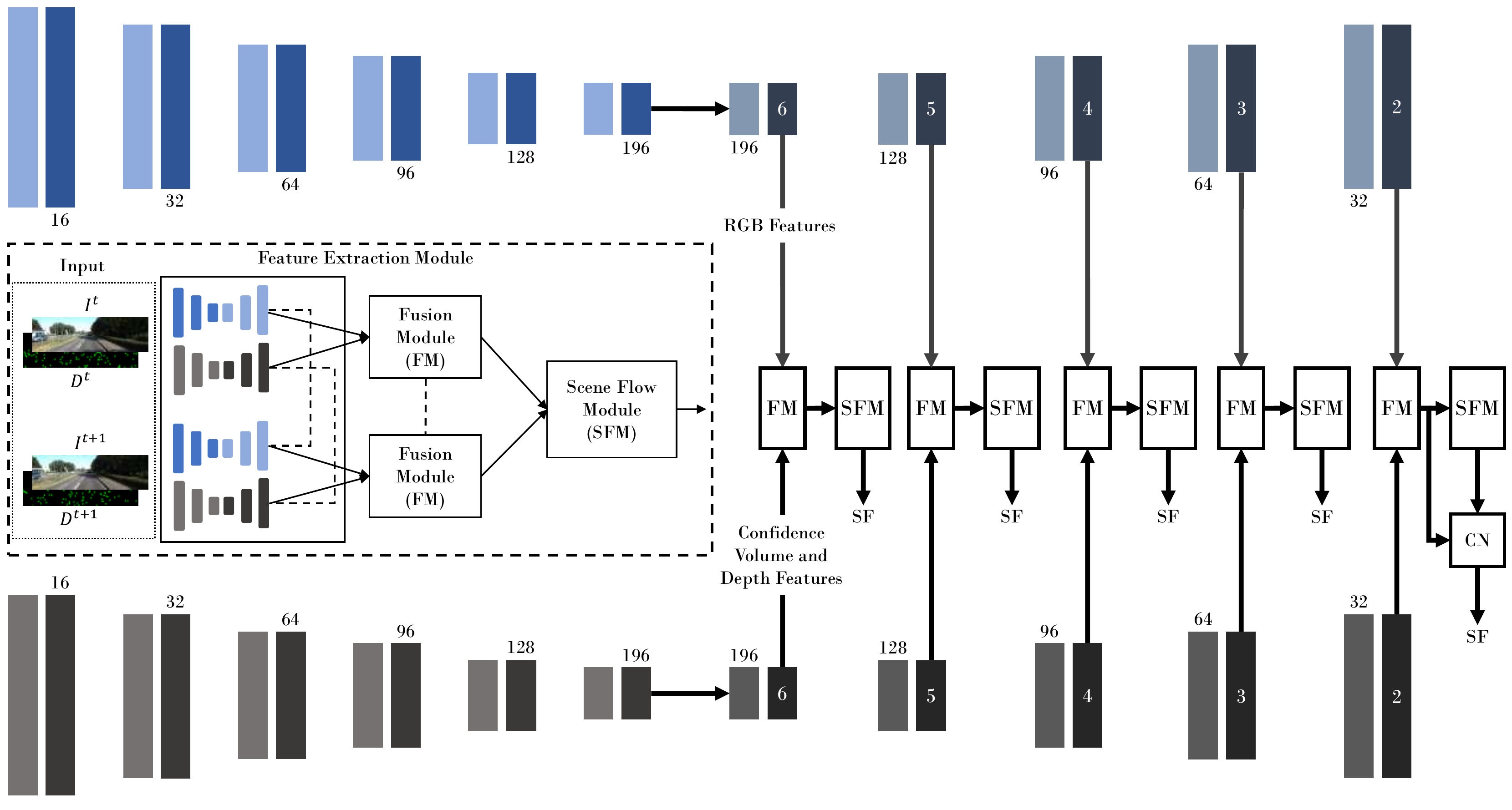}
    \caption{Detailed architecture of our \name{}. Residual connections from the feature pyramid are omitted for clarity. RGB features from the RGB pyramid and confidence volume along with the depth features from the confidence pyramid are sent to Fusion Module (FM). The fused features are then sent to the Scene Flow Module (SFM). The numbers $2$ to $6$ denote the levels which are used for multi-scale prediction. The output of level $2$ is refined in the context network (CN) to form the final scene flow. The number of channels are same for blocks of similar size.}
    \label{figure:deeplidarflow}
\end{figure*}

%% file: experiments.tex
\section{Experiments}
We conduct several experiments to verify the results of our \name{}. Firstly, we verify our design decisions. Secondly, we show the robustness of our \name{} compared to state-of-the approaches and we investigate finally the performance over different sparsity levels of the LiDAR, compared to a conventional approach.
  
\subsection{Data Sets and Evaluation Metrics}
\textbf{Data Sets: } Since the prime objective of this work is to predict scene flow for autonomous systems, KITTI \cite{geiger2012we} is a direct choice. The train set of KITTI consists of 200 consecutive frames with ground truth of scene flow aligned to a reference frame at time step $t$ and represented as optical flow components (Fl) and disparity maps (D0) and (D1) for a consecutive time steps $t$ and $t+1$ respectively. The disparity maps are generated using a high resolution LiDAR sensor and projected into image coordinate as disparity maps by using the calibration parameters. We de-warp the LiDAR frame of time step $t+1$ to mimic the real capture of the second LiDAR frame same as in \cite{battrawy2019lidar}. However, the established LiDAR frames are insufficient for training a deep neural network, hence, for all our experiments, we first pre-train our \name{} on FlyingThings3D (FT3D) \cite{mayer2016large} and then fine tune it on KITTI \cite{geiger2012we}. We split the train set of KITTI into training and validation splits and we conduct the same validation frames to the stat-of-the art methods mentioned in Section \ref {Robustness_Comparison} for a fair comparison. 

\textbf{Evaluation Metrics: } We split our metrics into two categories: 
\textit{Dense scene flow evaluation} -- We compute the average KITTI outlier rate for scene flow (SF) and its components (i.e. (D0), (D1) and (Fl)) over all pixels for which the endpoint error is $>$ 3 pixels and the relative error is $>$ 5~\% compared to the ground truth. Additionally, the euclidean distance (the endpoint error (SF-EPE)) is computed over all scene flow components.   
\textit{Sparse scene flow evaluation} -- Same thresholds as in dense evaluation are used to compute the outlier rate of optical (Fl) and we also compute the endpoint error for 2D optical flow (Fl-EPE) but only for the sparse input. In addition to these metrics, we consider 3D space metrics by projecting the input points and the measured displacement of scene flow as well as the ground truth into 3D Cartesian points. The average outlier rate of scene flow (SF-3D) is computed over all 3D input points whose euclidean distance to ground truth (endpoint error (SF-EPE-3D)) is $>$ 0.3 meters and the relative error is $>$ 10~\%.

\subsection{Implementation and Training}
Since FT3D and KITTI data sets have dense disparity maps, we use a uniform random sampling strategy to sample disparity points. Most real LiDAR sensors have some inherent amount of noise and to mimic this characteristic, some noise is simulated into the sampled depth points during training and fine tuning. Additionally, we apply the same data augmentation as in \cite{dosovitskiy2015flownet} for the RGB input. For training our architecture, we use the hyper-parameters and a multi-level losses as in \cite{saxena2019pwoc}. Noticeable, when trained with a fixed number of LiDAR points, the accuracy of \name{} is deteriorated a lot when testing with differently dense LiDAR input which is not a desirable characteristic. To overcome this problem, we generalize our model across different sparsity levels of LiDAR (i.e. resolutions) by varying the number of LiDAR samples on the fly (points are varied from 0.2\% to 20\% of full density) for both pre-training and fine-tuning. Using this strategy, our \name{} is able to achieve an almost constant accuracy for a wide range of LiDAR points.

\subsection{Design Decisions}
\label{design_decisions}
Our \name{} handles simultaneously the densification of LiDAR features to predict the dense scene flow. In this context, there are several questions that may come up, \textit{do we really need confidence convolution? Do we need to concatenate the confidences during fusion?}. To answer these questions, we conduct an experiment where the LiDAR pyramid uses regular convolution layers (i.e. no confidence convolution) and the results with this case are worse than when using confidence convolutions as shown in Table \ref{ablation}. In the fusion module, confidence volumes are concatenated to the RGB and LiDAR features. These confidence volumes act as meta-guidance to the fusion module, this also improves the final results as shown in Table \ref{ablation}.

\begin{table}[b]
	\centering
	\caption{Ablation study on various design choices for our \name{}. We test all variants 5000 points as LiDAR input on our test splits of FT3D and KITTI.}
	\resizebox{\linewidth}{!}
	{
	\begin{tabular}{c|c|cc|cc}
		\multirow{2}{*}{\makecell{\textbf{Confidence} \\ \textbf{Convolution}}} & \multirow{2}{*}{\makecell{\textbf{Confidence} \\ \textbf{Concatenation}}}& \multicolumn{2}{c|}{\textbf{FT3D \cite{mayer2016large}}} & \multicolumn{2}{c}{\textbf{KITTI \cite{geiger2012we}}} \Bstrut\\ 
		%\cline{3-6}
		& & \textbf{SF} & \textbf{SF-EPE} & \textbf{SF} & \textbf{SF-EPE}\Tstrut\Bstrut\\
		\hline 
		\xmark & \xmark & 30.97	& 8.77 & 16.33 & 3.75\Tstrut\\
		\cmark & \xmark & 21.83	& 8.70 & 16.31 & 4.05\Tstrut\\
		\cmark & \cmark & \textbf{20.20} & \textbf{7.64} & \textbf{13.77} & \textbf{3.67}\Tstrut\\
	\end{tabular}
	}
	\label{ablation}
\end{table}

\subsection{Robustness and Comparison to State-of-the-Art} \label{Robustness_Comparison}
Since our \name{} claims that a fusion approach can yield robust estimates as compared to image-only and LiDAR-only approaches, we compare its performance to several state-of-the-art methods which utilize different input modalities. We verify the run time in milliseconds (ms) of our method compared to other methods using a GeForce GTX 1080 Ti.

One of the main advantages of using LiDAR as an input for scene flow methods is its robustness. Image-based methods rely heavily on the quality of the image and hence often fail in regions of the image which contain mirror-like reflections, ill-conditioned environment etc. For the qualitative results, we visualize multiple examples in Fig. \ref{robustness} to verify the robustness, the strong localization and the superior performance of our \name{} compared to image-only and LiDAR-only methods. 

\textbf{Comparison with Image-only Method:}
Our \name{} uses concepts like pyramids, warping, occlusion and cost volume. PWOC-3D \cite{saxena2019pwoc} also uses similar concepts but with a pair of stereo images as an input. Our \name{} is able to obtain good performance for a wide range of sparsity levels with an optimum of just 5000 LiDAR points. For this input density, our \name{} is compared to PWOC-3D on KITTI \cite{geiger2012we} and FT3D \cite{mayer2016large} as shown in Table \ref{tab:comparison_results}. Moreover, we visualize the robustness and the localization in three examples as shown in Fig. \ref{robustness_a}. These examples present occlusions, reflective surfaces and shadows which are often challenging examples for any of image-only approaches. In these areas, our \name{} has the capability to resolve them and to result in more accurate scene flow estimations.

\textbf{Comparison with LiDAR-only Method:}
HPLFlowNet \cite{gu2019hplflownet} utilizes LiDAR scans represented as 3D Cartesian coordinates (i.e. a point cloud) at two time steps to estimate scene flow. Since they use sparse points, we perform a sparse evaluation on KITTI using 8192 of LiDAR points (proposed sparse level in HPLFlowNet) to compare between HPLFlowNet and our \name{}. Since HPLFlowNet is originally evaluated without including ground surface, we present the results for HPLFlowNet once by excluding the ground surface and once with the ground surface included. However, we include the ground surface in our \name{} but evaluate the same sparse locations of LiDAR measurements as in HPLFlowNet only. Note that \name{} produces a dense result (w.r.t the image resolution) independent of the input density. Our \name{} outperforms comfortably HPLFlowNet over (Fl) and (Fl-EPE) metrics and achieves comparable results to HPLFlowNet in terms of the 3D metrics (SF-3D and SF-EPE-3D) as shown in Table \ref{tab:sparse_eval}. The qualitative results show as well our superior accuracy compared to \cite{gu2019hplflownet} (see Fig. \ref{robustness_b}).

\input{robustness_figures}

\input{tables}

\begin{figure}
	\centering
	\includegraphics[width=\linewidth]{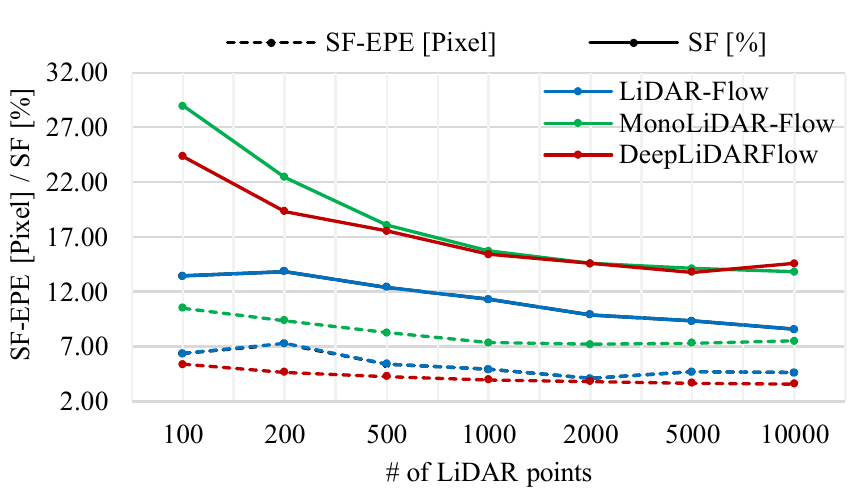}
	\caption{Our \name{} comparison against other fusion-based approaches in terms of scene flow outliers and scene flow endpoint error with varying number of LiDAR points. Dotted lines show the trend for endpoint error (SF-EPE) in pixels and the solid lines represent the outlier rate (SF) in percent. LiDAR-Flow \cite{battrawy2019lidar} outperforms all methods in terms of outlier rate (SF) since it exploits both LiDAR and RGB information in a stereo setup. Our \name{} performs better than its direct competitor MonoLiDAR-Flow for very sparse input. In terms of endpoint error, our \name{} comfortably outperforms all other methods.}
	\label{end_point_error}
\end{figure}

\textbf{Comparison with Image plus LiDAR Method:}
The only other method available in literature which uses the fusion of LiDAR and RGB images is LiDAR-Flow \cite{battrawy2019lidar} but it operates in a stereo setup and utilizes both stereo images and the corresponding LiDAR measurements. Therefore, it has much more information given as input than our \name{} which uses only monocular images and the corresponding LiDAR measurements. For having a fair comparison, we adopt a monocular setup with LiDAR measurements (called MonoLiDAR-Flow). To this end, we firstly densify the sparse LiDAR input by using the edge-preserving interpolation algorithm described in \cite{battrawy2019lidar}.
Secondly, we dissolve the stereo images in LiDAR-Flow pipeline \cite{battrawy2019lidar} to adopt only monocular setup with LiDAR.
The chart in Fig. \ref{end_point_error} compares MonoLiDAR-Flow, LiDAR-Flow and our \name{} with varying LiDAR densities on KITTI. LiDAR-Flow outperforms MonoLiDAR-Flow and our \name{} in terms of scene flow outliers (SF [\%]), a probable reason being the large amount of extra information it has due to the presence of a second camera view. Our \name{} outperforms MonoLiDAR-Flow for very sparse inputs, even for denser inputs; our \name{} results in an equivalent outliers rate compared to MonoLiDAR-Flow but our \name{} operates at a much higher speed than MonoLiDAR-Flow. In terms of (SF-EPE [px]), our \name{} outperforms LiDAR-Flow and MonoLiDAR-Flow consistently for all input densities. Table \ref{tab:comparison_results} presents a detailed comparison of these methods with all other metrics, when evaluated with a constant number of points (5000 points). Our \name{} performs as good as these methods (and better on several metrics) while operating at a much higher speed. As qualitative comparison, we visualize an example in Fig. \ref{teaser} which shows strong localization and robust scene flow estimation compared to MonoLiDAR-Flow approach using $\sim$100 points.

%% file: robustness_figures.tex
\begin{figure*}[t]
    \begin{subfigure}[c]{\textwidth}
    \hspace{0.04\textwidth}
    \begin{subfigure}[c]{0.15\textwidth}
        \centering
        \small
        Image-only
    \end{subfigure}
    \begin{subfigure}[c]{0.15\textwidth}
        \centering
        \small
        DeepLiDARFlow
    \end{subfigure}
    \begin{subfigure}[c]{0.15\textwidth}
        \centering
        \small
        Image-only
    \end{subfigure}
    \begin{subfigure}[c]{0.15\textwidth}
        \centering
        \small
        DeepLiDARFlow
    \end{subfigure}
    \begin{subfigure}[c]{0.15\textwidth}
        \centering
        \small
        Image-only
    \end{subfigure}
    \begin{subfigure}[c]{0.15\textwidth}
        \centering
        \small
        DeepLiDARFlow
    \end{subfigure}
    
    \vspace{1mm}
    \begin{subfigure}[c]{0.04\textwidth}
        \centering
       	\begin{turn}{90}\textit{\scriptsize Input}\end{turn}
    \end{subfigure}
    \begin{subfigure}[c]{0.15\textwidth}
        \centering
        \includegraphics[width =\linewidth, height=1cm]{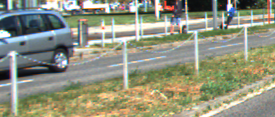}
    \end{subfigure}
    \begin{subfigure}[c]{0.15\textwidth}
        \centering
        \includegraphics[width =\linewidth, height=1cm]{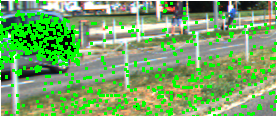}
    \end{subfigure}
    \begin{subfigure}[c]{0.15\textwidth}
        \centering
        \includegraphics[width =\linewidth, height=1cm]{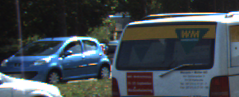}
    \end{subfigure}
    \begin{subfigure}[c]{0.15\textwidth}
        \centering
        \includegraphics[width =\linewidth, height=1cm]{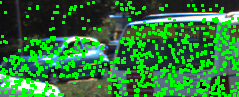}
    \end{subfigure}
    \begin{subfigure}[c]{0.15\textwidth}
        \centering
        \includegraphics[width =\linewidth, height=1cm]{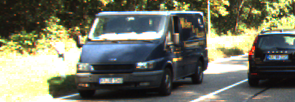}
    \end{subfigure}
    \begin{subfigure}[c]{0.15\textwidth}
        \centering
        \includegraphics[width =\linewidth, height=1cm]{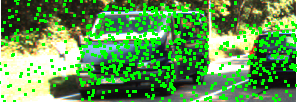}
    \end{subfigure}
    
    \vspace{1mm}
    \begin{subfigure}[c]{0.04\textwidth}
        \centering
       	\begin{turn}{90}\textit{\scriptsize D0}\end{turn}
    \end{subfigure}
    \begin{subfigure}[c]{0.15\textwidth}
        \centering
        \includegraphics[width =\linewidth, height=1cm]{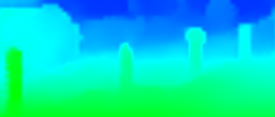}
    \end{subfigure}
    \begin{subfigure}[c]{0.15\textwidth}
        \centering
        \includegraphics[width =\linewidth, height=1cm]{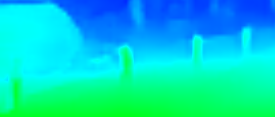}
    \end{subfigure}
    \begin{subfigure}[c]{0.15\textwidth}
        \centering
        \includegraphics[width =\linewidth, height=1cm]{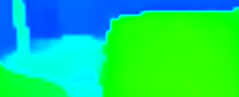}
    \end{subfigure}
    \begin{subfigure}[c]{0.15\textwidth}
        \centering
        \includegraphics[width =\linewidth, height=1cm]{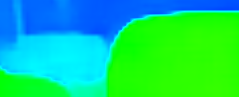}
    \end{subfigure}
    \begin{subfigure}[c]{0.15\textwidth}
        \centering
        \includegraphics[width =\linewidth, height=1cm]{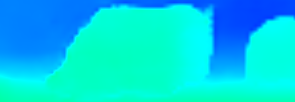}
    \end{subfigure}
    \begin{subfigure}[c]{0.15\textwidth}
        \centering
        \includegraphics[width =\linewidth, height=1cm]{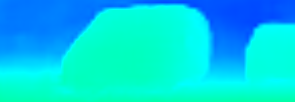}
    \end{subfigure}
    
    \vspace{1mm}
    \begin{subfigure}[c]{0.04\textwidth}
        \centering
       	\begin{turn}{90}\textit{\scriptsize Fl}\end{turn}
    \end{subfigure}
    \begin{subfigure}[c]{0.15\textwidth}
        \centering
        \includegraphics[width =\linewidth, height=1cm]{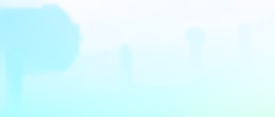}
    \end{subfigure}
    \begin{subfigure}[c]{0.15\textwidth}
        \centering
        \includegraphics[width =\linewidth, height=1cm]{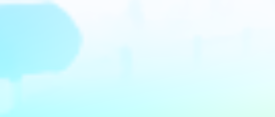}
    \end{subfigure}
    \begin{subfigure}[c]{0.15\textwidth}
        \centering
        \includegraphics[width =\linewidth ,height=1cm]{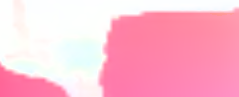}
    \end{subfigure}
    \begin{subfigure}[c]{0.15\textwidth}
        \centering
        \includegraphics[width =\linewidth, height=1cm]{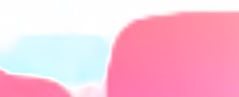}
    \end{subfigure}
    \begin{subfigure}[c]{0.15\textwidth}
        \centering
        \includegraphics[width =\linewidth, height=1cm]{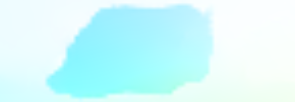}
    \end{subfigure}
    \begin{subfigure}[c]{0.15\textwidth}
        \centering
        \includegraphics[width =\linewidth, height=1cm]{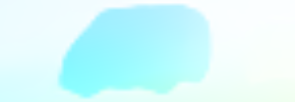}
    \end{subfigure}
    
    \vspace{1mm}
    \begin{subfigure}[c]{0.04\textwidth}
        \centering
       	\begin{turn}{90}\textit{\scriptsize Error Map}\end{turn}
    \end{subfigure}
    \begin{subfigure}[c]{0.15\textwidth}
        \centering
        \includegraphics[width =\linewidth, height=1cm]{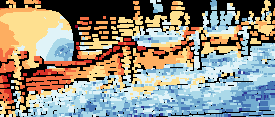}
    \end{subfigure}
    \begin{subfigure}[c]{0.15\textwidth}
        \centering
        \includegraphics[width =\linewidth, height=1cm]{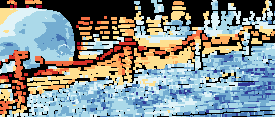}
    \end{subfigure}
    \begin{subfigure}[c]{0.15\textwidth}
        \centering
        \includegraphics[width =\linewidth, height=1cm]{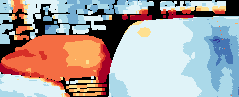}
    \end{subfigure}
    \begin{subfigure}[c]{0.15\textwidth}
        \centering
        \includegraphics[width =\linewidth, height=1cm]{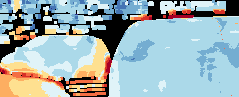}
    \end{subfigure}
    \begin{subfigure}[c]{0.15\textwidth}
        \centering
        \includegraphics[width =\linewidth, height=1cm]{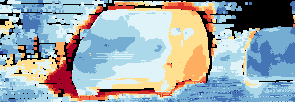}
    \end{subfigure}
    \begin{subfigure}[c]{0.15\textwidth}
        \centering
        \includegraphics[width =\linewidth, height=1cm]{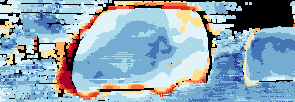}
    \end{subfigure}
    
    \vspace{1mm}
    \begin{subfigure}[c]{0.04\textwidth}
		\raggedright \scriptsize EPE
    \end{subfigure}
    \begin{subfigure}[c]{0.935\textwidth}
    	\includegraphics[width=\linewidth]{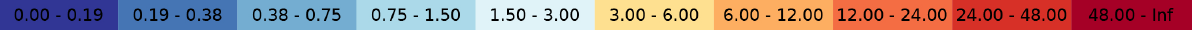}
    \end{subfigure}
    \caption{Image-only vs. our \name{}}
    \label{robustness_a}
    \end{subfigure}

    \vspace{3mm}
    \begin{subfigure}[c]{\textwidth}
    \hspace{0.04\textwidth}
    \begin{subfigure}[c]{0.15\textwidth}
        \centering
        \small
        LiDAR-only
    \end{subfigure}
    \begin{subfigure}[c]{0.15\textwidth}
        \centering
        \small
        \name{}
    \end{subfigure}
    \begin{subfigure}[c]{0.15\textwidth}
        \centering
        \small
        LiDAR-only
    \end{subfigure}
    \begin{subfigure}[c]{0.15\textwidth}
        \centering
        \small
        \name{}
    \end{subfigure}
    \begin{subfigure}[c]{0.15\textwidth}
        \centering
        \small
        LiDAR-only
    \end{subfigure}
    \begin{subfigure}[c]{0.15\textwidth}
        \centering
        \small
        \name{}
    \end{subfigure}
    
    \vspace{1mm}
    \begin{subfigure}[c]{0.04\textwidth}
        \centering
       	\begin{turn}{90}\textit{\scriptsize Input}\end{turn}
    \end{subfigure}
    \begin{subfigure}[c]{0.15\textwidth}
        \centering
        \includegraphics[width =\linewidth, height=1cm]{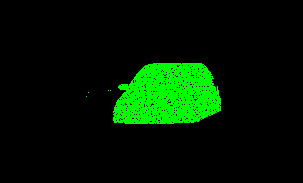}
    \end{subfigure}
    \begin{subfigure}[c]{0.15\textwidth}
        \centering
        \includegraphics[width =\linewidth, height=1cm]{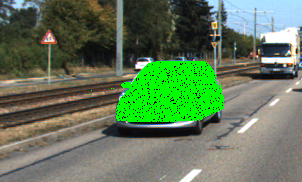}
    \end{subfigure}
    \begin{subfigure}[c]{0.15\textwidth}
        \centering
        \includegraphics[width =\linewidth, height=1cm]{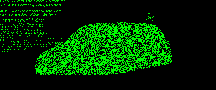}
    \end{subfigure}
    \begin{subfigure}[c]{0.15\textwidth}
        \centering
        \includegraphics[width =\linewidth, height=1cm]{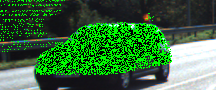}
    \end{subfigure}
    \begin{subfigure}[c]{0.15\textwidth}
        \centering
        \includegraphics[width =\linewidth, height=1cm]{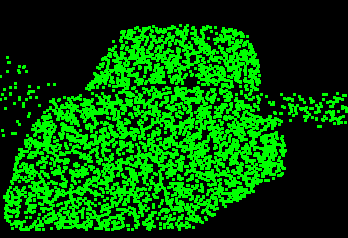}
    \end{subfigure}
    \begin{subfigure}[c]{0.15\textwidth}
        \centering
        \includegraphics[width =\linewidth, height=1cm]{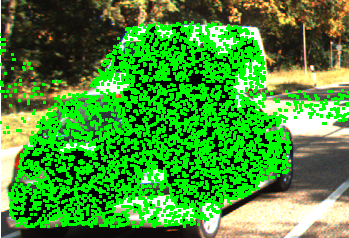}
    \end{subfigure}
    
    \vspace{1mm}
    \begin{subfigure}[c]{0.04\textwidth}
        \centering
       	\begin{turn}{90}\textit{\scriptsize Fl}\end{turn}
    \end{subfigure}
    \begin{subfigure}[c]{0.15\textwidth}
        \centering
        \includegraphics[width =\linewidth, height=1cm]{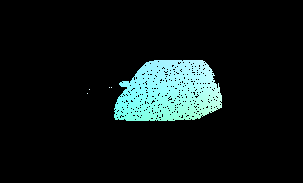}
    \end{subfigure}
    \begin{subfigure}[c]{0.15\textwidth}
        \centering
        \includegraphics[width =\linewidth, height=1cm]{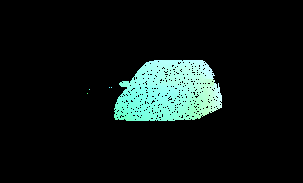}
    \end{subfigure}
    \begin{subfigure}[c]{0.15\textwidth}
        \centering
        \includegraphics[width =\linewidth ,height=1cm]{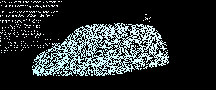}
    \end{subfigure}
    \begin{subfigure}[c]{0.15\textwidth}
        \centering
        \includegraphics[width =\linewidth, height=1cm]{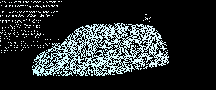}
    \end{subfigure}
    \begin{subfigure}[c]{0.15\textwidth}
        \centering
        \includegraphics[width =\linewidth, height=1cm]{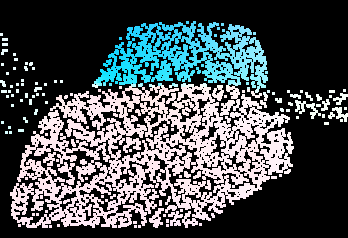}
    \end{subfigure}
    \begin{subfigure}[c]{0.15\textwidth}
        \centering
        \includegraphics[width =\linewidth, height=1cm]{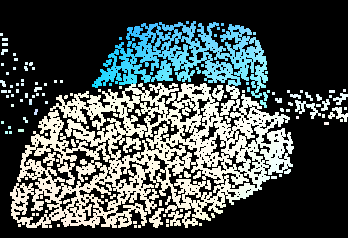}
    \end{subfigure}
    
    \vspace{1mm}
    \begin{subfigure}[c]{0.04\textwidth}
        \centering
       	\begin{turn}{90}\textit{\scriptsize Error Map}\end{turn}
    \end{subfigure}
    \begin{subfigure}[c]{0.15\textwidth}
        \centering
        \includegraphics[width =\linewidth, height=1cm]{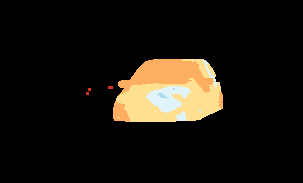}
    \end{subfigure}
    \begin{subfigure}[c]{0.15\textwidth}
        \centering
        \includegraphics[width =\linewidth, height=1cm]{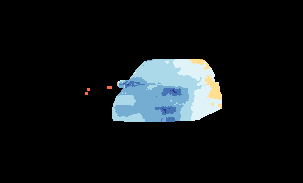}
    \end{subfigure}
    \begin{subfigure}[c]{0.15\textwidth}
        \centering
        \includegraphics[width =\linewidth, height=1cm]{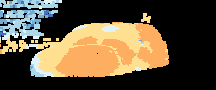}
    \end{subfigure}
    \begin{subfigure}[c]{0.15\textwidth}
        \centering
        \includegraphics[width =\linewidth, height=1cm]{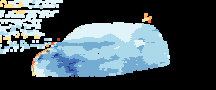}
    \end{subfigure}
    \begin{subfigure}[c]{0.15\textwidth}
        \centering
        \includegraphics[width =\linewidth, height=1cm]{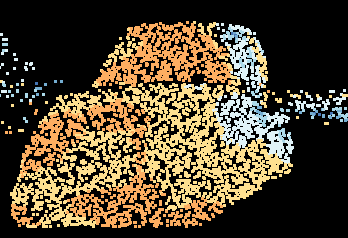}
    \end{subfigure}
    \begin{subfigure}[c]{0.15\textwidth}
        \centering
        \includegraphics[width =\linewidth, height=1cm]{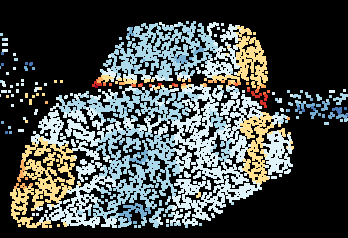}
    \end{subfigure}
    
    \vspace{1mm}
    \begin{subfigure}[c]{0.04\textwidth}
		\raggedright \scriptsize EPE
    \end{subfigure}
    \begin{subfigure}[c]{0.935\textwidth}
    	\includegraphics[width=\linewidth]{Graphics/KITTI_errorcolors}
    \end{subfigure}
    \caption{LiDAR-only vs. our \name{} (ground surface removed in outputs and error maps)}
    \label{robustness_b}
    \end{subfigure}
    \caption{Our \name{} presents high robustness using LiDAR features and the rich textural information of RGB features. Here we compare some results from \name{} against an image-only approach \cite{saxena2019pwoc} and a LiDAR-only approach \cite{gu2019hplflownet}. \name{} shows superior performance in regions of bad illumination compared to the image-only approach, and overcomes the problem of unstructured point clouds yielding a result of much higher resolution, compared to LiDAR-only.}
    \label{robustness}
\end{figure*}

%% file: tables.tex
\begin{table*}
    \centering
    \caption{Comparison of scene flow results for PWOC-3D \cite{saxena2019pwoc}, LiDAR-Flow \cite{battrawy2019lidar}, MonoLiDAR-Flow (monocular version of LiDAR-Flow), and \name{} on the test splits of KITTI \cite{geiger2012we} and FT3D \cite{mayer2016large}. LiDAR methods are evaluated with an input of 5000 depth measurements.}
    \resizebox{\textwidth}{!}{
    \begin{tabular}{c|c|ccccc|ccccc|c}
        \multirow{2}{*}{\textbf{Method}} & \multirow{2}{*}{\textbf{Modality}} & \multicolumn{5}{c|}{\textbf{KITTI} \cite{geiger2012we}} & \multicolumn{5}{c|}{\textbf{FT3D} \cite{mayer2016large}} & \multirow{2}{*}{\textbf{Time (ms)}}\\
         & & \textbf{D0} & \textbf{D1} & \textbf{Fl} & \textbf{SF} & \textbf{SF-EPE} & \textbf{D0} & \textbf{D1} & \textbf{Fl} & \textbf{SF} & \textbf{SF-EPE} & \Tstrut\Bstrut\\
         \hline
         PWOC-3D \cite{saxena2019pwoc} & Stereo-Only  & 4.07	&6.1	&10.29	&12.24	& \textbf{3.15} & 8.04 & 9.30& 16.64&19.30 & \textbf{6.97} & \textbf{130} \Tstrut\\
          LiDAR-Flow \cite{battrawy2019lidar} & Stereo + LiDAR & \textbf{2.30}	& \textbf{5.03}	& \textbf{8.46}	& \textbf{9.33}	& 4.67 & \textbf{3.69} & \textbf{6.48} &\textbf{15.10} & \textbf{16.00}	&29.97 & 65900 \Tstrut\Bstrut\\
          \hline
         MonoLiDAR-Flow &  Monocular + LiDAR & \textbf{2.10}	& \textbf{6.55}	& 13.37	&14.11	& 7.31 & \textbf{4.04}	& \textbf{5.80}	& \textbf{15.04}	& \textbf{16.02}	&24.29 & 34700 \Tstrut\\
         Our \name{} & Monocular + LiDAR & 4.18	&7.33	& \textbf{11.26}	&\textbf{13.77}	& \textbf{3.64} & 6.13	&7.75	&18.51	&20.34	&\textbf{6.87} & \textbf{310} \Tstrut\\
    \end{tabular}
    }
    \label{tab:comparison_results}
\end{table*}

\begin{table*}
	\centering
	\caption{Sparse evaluation of \name{} and HPLFlowNet \cite{gu2019hplflownet} with and without ground surface on KITTI. When grounds are included, our \name{} outperforms HPLFlowNet significantly over all terms. HPLFlowNet is able to outperform our \name{} only in terms of 3D metrics (i.e. SF-EPE-3D and SF-3D) when the ground surface is removed. Even then, our \name{} has better performance in terms of optical flow estimation.}
	\begin{tabular}{c|c|c|c|cccc|c}
		\multirow{2}{*}{\textbf{Method}} & \multirow{2}{*}{\textbf{Modality}} & \multirow{2}{*}{\makecell{\textbf{Ground}\\\textbf{Surface}}} & \multirow{2}{*}{\makecell{\textbf{Output}\\\textbf{Density}}} & \multicolumn{4}{c|}{\textbf{KITTI} \cite{geiger2012we}} & \multirow{2}{*}{\textbf{Time (ms)}}\\
	& & & & \textbf{Fl-EPE} & \textbf{Fl} & \textbf{SF-EPE-3D} & \textbf{SF-3D}\Tstrut\Bstrut\\
	\hline
	HPLFlowNet \cite{gu2019hplflownet} & LiDAR-Only & excluded & $\sim$~2~\% & 5.94&47.54& \textbf{0.14} & \textbf{12.79} & \textbf{301}\Tstrut\\
	HPLFlowNet \cite{gu2019hplflownet} & LiDAR-Only & included & $\sim$~2~\% & 9.77	&71.62	&0.27	&33.84 & \textbf{301}\Tstrut\\
	Our \name{} & Monocular + LiDAR & included & \textbf{100.0~\%} & \textbf{2.89}	& \textbf{11.74}	& 0.15	&13.27 & 310 \Tstrut\\
	\end{tabular}
	\label{tab:sparse_eval}
\end{table*}